\documentclass{article}
\usepackage{neurips_2020, hyperref, amsfonts, amsmath, graphicx, lineno, xcolor, setspace}

\title{Deep Learning of the Evolution Operator Enables Forecasting of Out-of-Training Dynamics in Chaotic Systems}

\author{
Ira J. S. Shokar$^1$, Peter H. Haynes$^1$, Rich R. Kerswell$^1$   \\
$^1$Department of Applied Mathematics and Theoretical Physics, University of Cambridge,\\ Wilberforce Road, Cambridge, CB3 0WA, UK\\
}

\begin{document}
\begin{spacing}{1.15}
\maketitle


\begin{abstract}
We demonstrate that a deep learning emulator for chaotic systems can forecast phenomena absent from training data. Using the Kuramoto-Sivashinsky and beta-plane turbulence models, we evaluate the emulator through scenarios probing the fundamental phenomena of both systems: forecasting spontaneous relaminarisation, capturing initialisation of arbitrary chaotic states, zero-shot prediction of dynamics with parameter values outside of the training range, and characterisation of dynamical statistics from artificially restricted training datasets. Our results show that deep learning emulators can uncover emergent behaviours and rare events in complex systems by learning underlying mathematical rules, rather than merely mimicking observed patterns.
\end{abstract}


Chaotic systems are known to exhibit sensitivity to initial conditions, limiting long-term predictability. This unpredictability challenges the forecasting of large-scale behaviours and rare or extreme events, such as sudden regime shifts or extreme weather occurrences in the climate system, for example, which are critical for understanding the system's dynamics \cite{pred_chaos}. 

Recent advances in machine learning (ML) have shown significant promise in modelling chaotic dynamics \cite{pathek}. However, a fundamental limitation remains unclear: do these models inherently fail to predict beyond what is seen during training? This can manifest in two main ways: an inability to predict unseen dynamics at the same parameter values of the training set, for a given dynamical system, or the inability to make predictions into new parameter regimes beyond the training data (the 'extrapolation' problem \cite{physics_ML}). While the latter is unfortunate, the former is more critical as it calls into question whether an ML model can be used to predict extreme or rare events which are infrequent in or entirely absent in the training dataset \cite{fluids_ML_review, pedram, pred_lim, cautionary, extrap}. Understanding whether these limitations can be circumvented is crucial for unlocking the full potential of ML as a tool for understanding chaotic systems.

In this paper, we address both limitations by demonstrating that a deep learning emulator {\em can} successfully predict unobserved dynamics in two chaotic systems in and beyond the parameter regimes of the training data set. The chaotic systems considered are the Kuramoto-Sivashinsky (KS) equation \cite{unstableKS}, a well-studied canonical model of spatiotemporal chaos, and beta-plane turbulence \cite{beta_turb}, a useful simplest model for many aspects of flows in atmospheres and oceans. The emulator constructed accurately reproduces known behaviours and predicts phenomena not present in the training data, including relaminarisation events and initialisation dynamics. We believe that this is because the emulator constructs a sufficiently accurate approximation of the evolution operator ${\cal F}_{\Delta t}$ that advances the solutions of the underlying partial differential equation (PDE), $\partial_t u = F(u)$, over time $u_{t+\Delta t} = {\cal F}_{\Delta t} (u_t)$, without having to see all possible dynamical behaviours allowed in these systems, i.e. it is able to do more than just learn observed patterns.
%
%
\begin{figure*}[t!] 
    \centering
    \includegraphics[width=\textwidth]{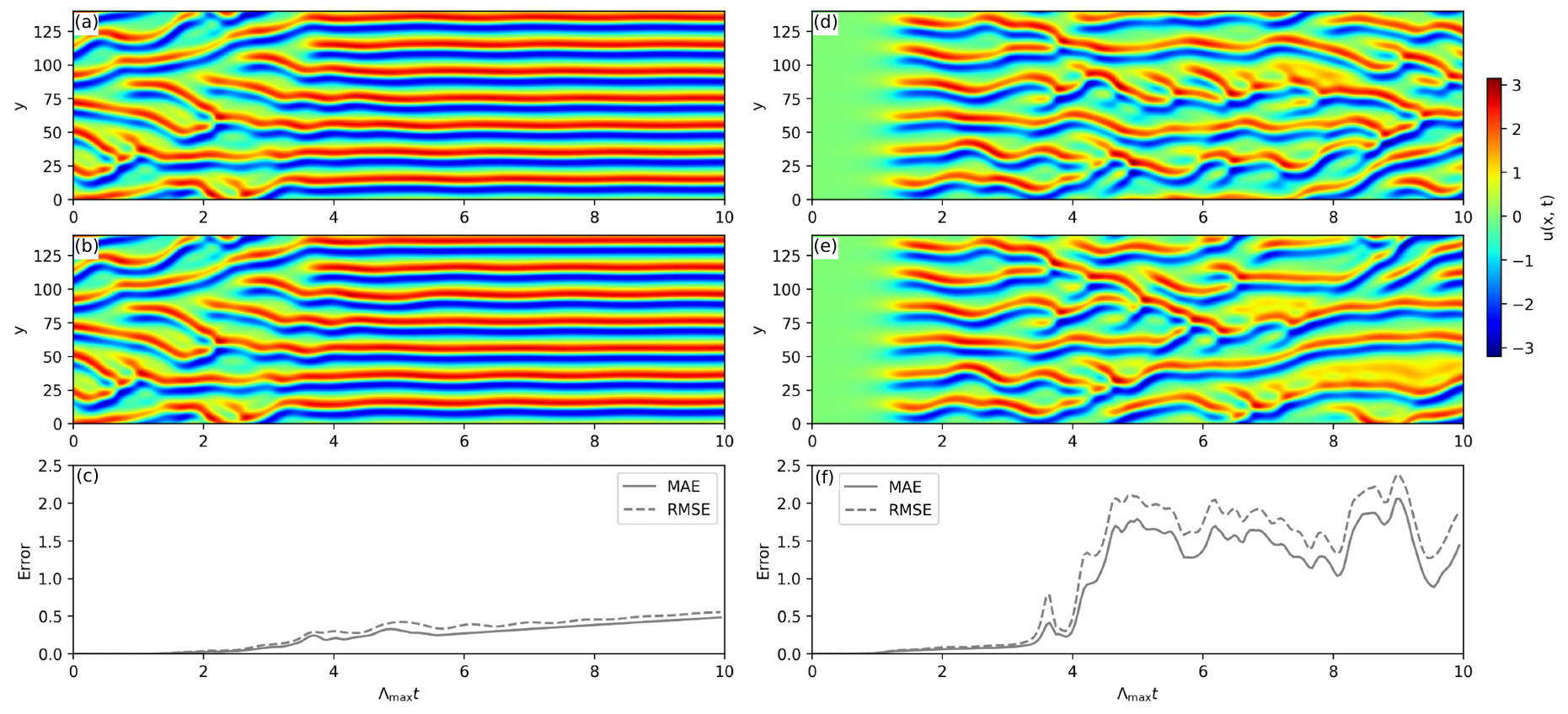}
    \caption{Prediction of out-of-training-distribution dynamics of the KS equation with $L=56$, where training data excludes relaminarisation events or warm-up dynamics.
    (a) Relaminarisation event observed in direct numerical simulation (DNS).
    (b) Neural network (NN) emulation predicts the relaminarisation event from identical initial conditions.
    (c) Mean Absolute Error (MAE) and Root Mean Squared Error (RMSE) between the numerical simulation (a) and NN prediction (b).
    (d) Initialisation dynamics of KS flow from DNS.
    (e) NN emulation forecasts the correct dynamics from the same initial conditions.
    (f) MAE and RMSE between the initialisation dynamics (d) and NN prediction (e).
    The largest Lyapunov exponent for $L=56$ is $\Lambda_{\text{max}} \approx 0.048$.}
\end{figure*}

Our first system is the Kuramoto-Sivashinsky (KS) equation, defined as $F(u) := -u \partial_y u - \partial_{yy}^2 u - \partial_{yyyy}^4 u$, where $u(y,t)$ is a scalar field on the periodic domain $[0, L)$. As the domain size $L$ increases, the system undergoes a series of bifurcations, leading to increasingly complex behaviour \cite{unstableKS}. Previous work on emulating the KS equation has been limited to fixed domain sizes \cite{KS0, KS1, KS2, Chen_2021} or requires \textit{ab initio} training for different $L$ values \cite{pathek}. Instead, we develop an emulator capable of handling varying spatial dimensions while learning parameter-dependent dynamics (see End Matter for details). This employs a transformer \cite{Attention} with local attention \cite{SASA} to capture spatial correlations across scales while efficiently handling inputs of any size. Restricting correlations to local windows imposes a physical locality bias natural to the KS equation and many other PDEs as well as reducing the memory overhead typically associated with transformers. The network is conditioned on parameter $L$ via adaptive layer normalisation \cite{FILM}, which modulates activations to learn parametric (here domain-size) dependence-facilitating generalisation across parameter regimes.

Direct numerical simulation (DNS) of the KS equation was performed on a GPU using FourierFlows \cite{FourierFlows} in Julia \cite{Julia}, employing a pseudo-spectral method with a fourth-order exponential time differencing Runge-Kutta solver, with $\delta t=2.5\times10^{-2}$, which is subsampled to $\Delta t=1$ for training. Initialised with small amplitude noise $u_{\text{initial}} \sim \mathcal{N}(0, 0.01)$, only dynamics after a warm-up phase are used as training data. The network is pretrained on a weakly chaotic regime in a small domain ($L=22$), where $y \in \mathbb{R}^{56}$ (the domain is discretized by 56 uniformly-spaced grid points) using 5,000 snapshots, allowing the bulk of the computation to be done efficiently. The model is then fine-tuned, in the chaotic regime, on larger domains $L in \{22, 36, 48, 64, 98, 128, 200\}$, with $y \in \mathbb{R}^{\{56, 90, 120, 160, 246, 320, 500\}}$ respectively to maintain the same grid spacing, and using 500 snapshots each to learn dependence on the parameter $L$. The optimisation problem solved in training is to minimise the Mean Squared Error $\text{MSE} = \frac{1}{L} \int_{0}^{L} \left|u_{t+\Delta t} - \mathcal{F}_{\Delta t} ( u_{t} )\right|^2 dx$ over the one step prediction, to obtain an approximation for $\mathcal{F}_{\Delta t}$. The emulation is then auto-regressive, advancing the system in time according to the rule $\tilde{u}_{t+\Delta t} = \mathcal{F}_{\Delta t} (\tilde{u}_{t})$. For evaluation we use both the Mean Absolute Error ($\text{MAE}(t) = \frac{1}{L} \int_{0}^{L} \left| u_t - \tilde{u}_t \right| dx$) and Root Mean Squared Error ($\text{RMSE} = \sqrt{\text{MSE}}$).

%
%
We first start with relaminarisation—the process by which chaotic dynamics are attracted to an equilibrium state or a travelling wave solution \cite{back}, see Fig.1 (a). Despite being trained exclusively on chaotic dynamics for $L \neq 56$, the emulator accurately forecasts both the onset and progression of a transition to laminar dynamics at $L = 56$: see  Fig.1 (b). It captures the precise timing of the transition and reproduces the final stable wave pattern, demonstrating capture of the system's stability properties and the ability to predict the fixed point. Relaminarisation represents a qualitatively distinct flow regime from any represented in the training data, and capturing this regime requires the model to learn the fundamental mathematical rules describing the physics rather than relying on pattern matching. Fig.1 (c) shows the MAE and RMSE between the neural network (NN) emulation and DNS. These measures grow slowly due to a latitudinal translation drift in the laminar solution (an equally valid solution).

%
%
The emulator also accurately captures the transient growth phase in the KS equation, where small-amplitude noise evolves into fully developed chaos, as shown in Fig.1(d,e). This transition is characterised by the formation of localised structures (`kinks')—steep gradients in the solution that separate distinct spatial regions, visualised here as red-blue pairs. Note that it is not just the development of finite amplitude chaotic patterns, but the precise spatial structure that is captured. This is particularly notable because the amplitude of the fluctuations in $u$ during the initialisation phase is over two orders of magnitude smaller than those in the training set, representing a clear case of out-of-distribution dynamics. Fig.1 (f) displays the MAE and RMSE, illustrating the emulator’s predictive accuracy over time following the transition to chaos. The NN emulation remains accurate for approximately four Lyapunov times before diverging to an alternative but equally plausible solution.

%
%
\begin{figure}[t!] 
\centering
\includegraphics[width=\textwidth]{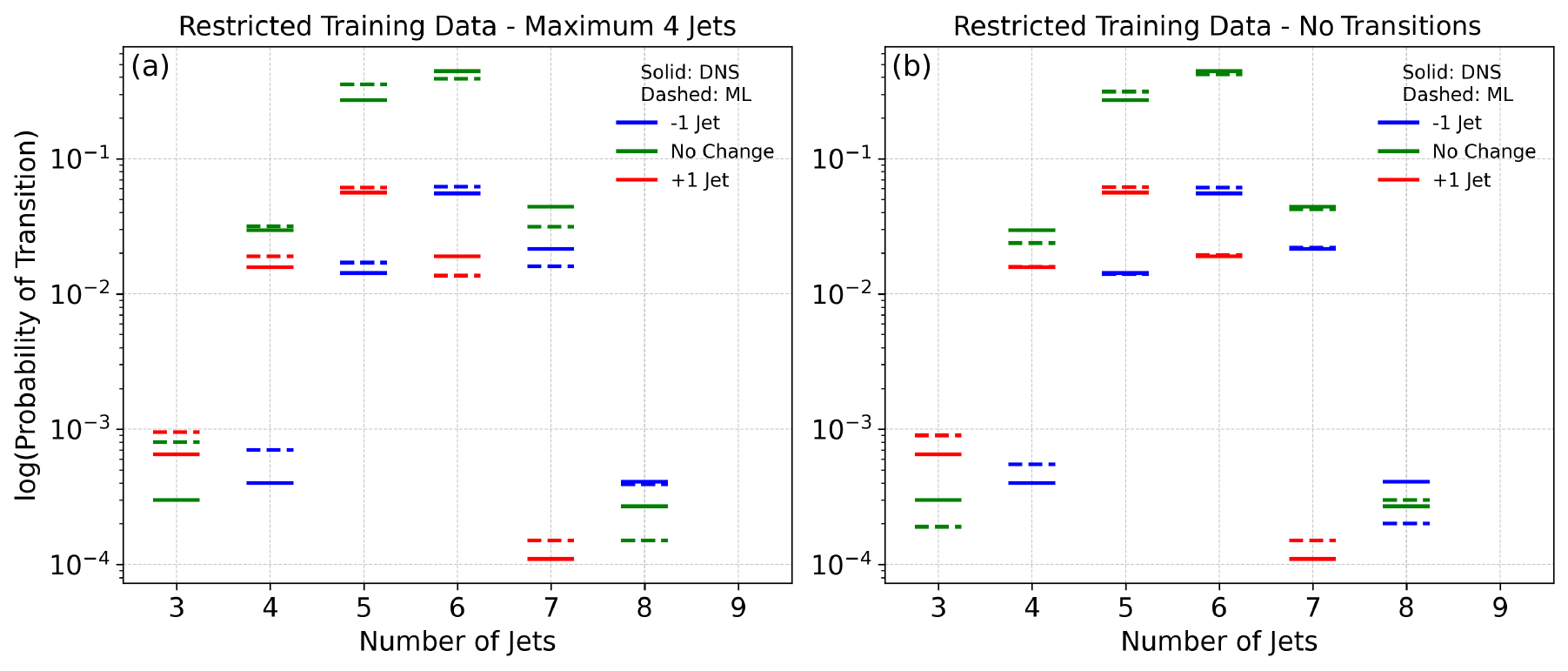}
\caption{Transition probabilities for kink count evolution in the KS equation, where the number of kinks (localised steep gradients) at time $t$ determines whether the count increases (red), decreases (blue), or remains constant (green) at time $t+\Delta t$.
(a) Comparison of kink count distributions for a restricted training dataset limited to four or fewer kinks, showing the distributions from full KS dynamics (solid lines) and the NN emulator trained on the restricted dataset (dashed lines).
(b) Same as (a), but with the restricted dataset excluding all transition events involving kink count changes.
}
\end{figure}

%
%
To further probe the model's capacity for out-of-distribution prediction, we train it on artificially restricted datasets that exclude key phenomena. The laminar dynamics (Fig.1 (c,d)) exhibit seven equally spaced kinks, as such, we restrict the training dataset to states with just four kinks or fewer to ensure it does not include any precursor dynamics to a relaminarisation event. As a result, transitions to states with five or more kinks are excluded. This is done by identifying the number of kinks in $u_t$ and tracking whether $u_{\Delta t}$ maintains the same kink count or undergoes a nucleation event (increase in kink count) or a coalescence event (decrease in kink count) (Fig.2 (a)). Using this restricted dataset (5,000 snapshots at $L=56$ in the chaotic regime), the emulator—trained exclusively on this limited domain without any pretraining or fine-tuning—accurately reproduces the full range of dynamical variability, including previously unseen 5+ kink states during a 10,000-step rollout. Furthermore, as shown in Fig.2 (a), the transition probabilities predicted by the NN closely align with those of the full DNS, despite the emulator being trained on a restricted regime.

We also generate a training dataset that excludes all transition events involving changes in the number of kinks, corresponding to the solid green lines in Fig.2 (b). While the network is exposed to the full range of kink numbers, it is not provided with any transition events or their precursory states (the 50 time snapshots preceding each transition are also excluded). To test whether the success of the emulator is simply due to providing a sufficiently large training dataset we only include 100 snapshots in the training dataset for this scenario. Despite these constraints on the training data, the emulator still successfully captures transitions between different kink states, with the predicted transition probabilities closely matching those of the full DNS. Moreover, in both cases, the models trained on domain-restricted datasets accurately predict both relaminarisation and initialisation dynamics for the test cases in Fig.1. This demonstrates that NNs are able to predict such out-of-training regime transitions without knowledge of events leading to such transitions \cite{tobias_ml}, a property previously considered beyond the reach of purely data-driven approaches \cite{GIN_2020}.

%
%
Finally, we assess the emulator's ability to extrapolate to $L=400$ after pretraining on $L=22$ and finetuning across smaller domains, $L \in \{22, 36, 48, 64, 98, 128, 200\}$—the same emulator used in Fig. 1. Fig. 3 shows that it successfully reproduces characteristic chaotic dynamics and delivers accurate short-term predictions, with RMSE reaching O(1) at $5/\Lambda_{\text{max}}$. This suggests that NN models may possess greater extrapolation capacity beyond their training distribution than traditionally assumed \cite{physics_ML}.

%
%
\begin{figure}[t!] 
\centering
\includegraphics[width=0.6\textwidth]{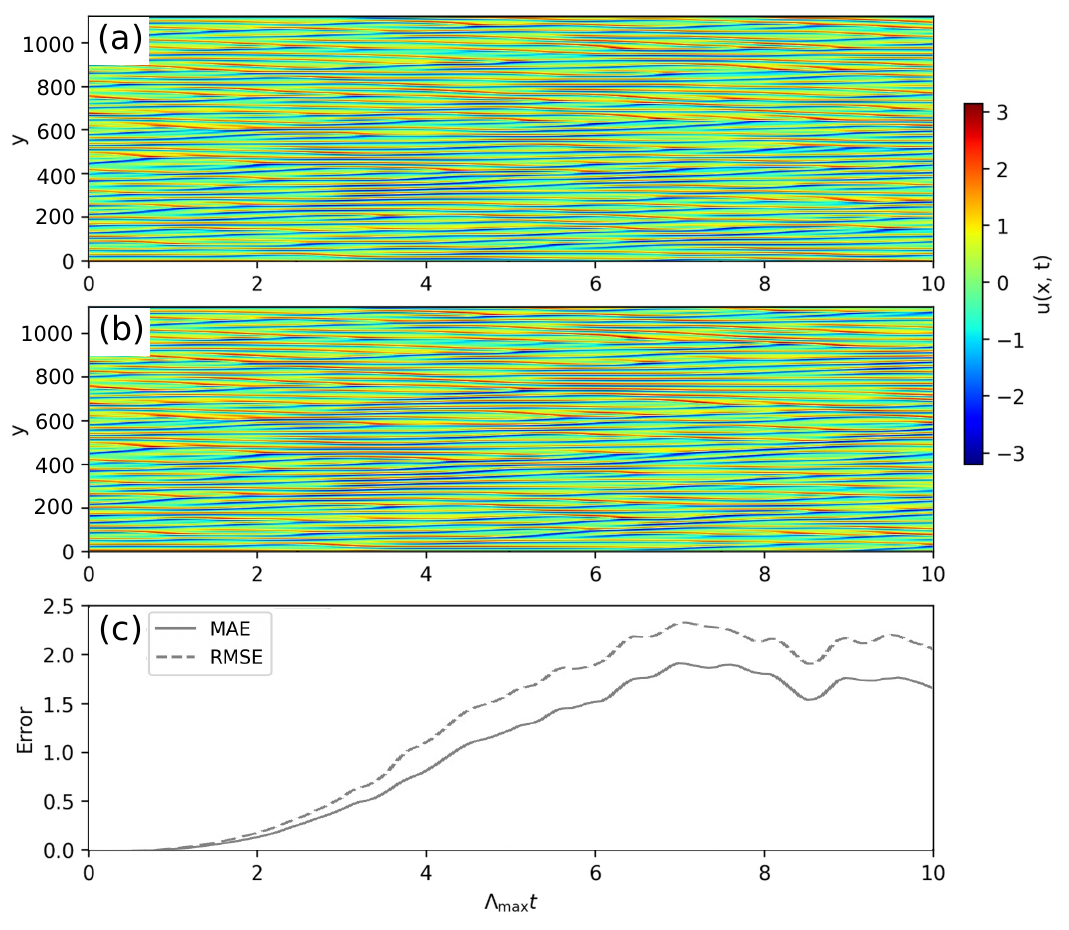}
\caption{Prediction of the Kuramoto-Sivashinsky (KS) equation dynamics with $L=400$, extrapolated beyond the training set.
(a) Numerical integration of the KS equation with $L=400$.
(b) NN emulation of the KS dynamics, with the emulator pretrained on data for $L=22$ and fine-tuned with datasets for $L \in \{48, 64, 96, 128, 164, 200\}$.
(c) MAE and RMSE comparing the numerical simulation in (a) to the NN prediction in (b).}
\end{figure}

%
%
It is important to emphasise that the emulator employed in this study is not the sole architecture capable of making out-of-training-set predictions. A fully connected NN with a single hidden layer can reproduce the results in Figs. 1 and 2 when trained on chaotic dynamics at $L=56$, but its fixed input size prevents generalisation to other $L$. Our attention-based emulator was chosen over alternatives like the Fourier Neural Operator (FNO) \cite{FNO} due to its superior extrapolation across parameter space, particularly for different geometries. While the FNO can generalise to varying parameters via adaptive layer normalisation, it struggles to capture both small and large $L$ simultaneously due to a fundamental scaling limitation: the number of resolvable wavenumbers is constrained by the smallest domain, restricting its ability to model larger domains with an increased number of active modes (see End Matter). In contrast, our emulator scales effectively across both small and large domains, overcoming these limitations.

%
%
%
%
\begin{figure}[t!] 
\centering
\includegraphics[width=0.6\textwidth]{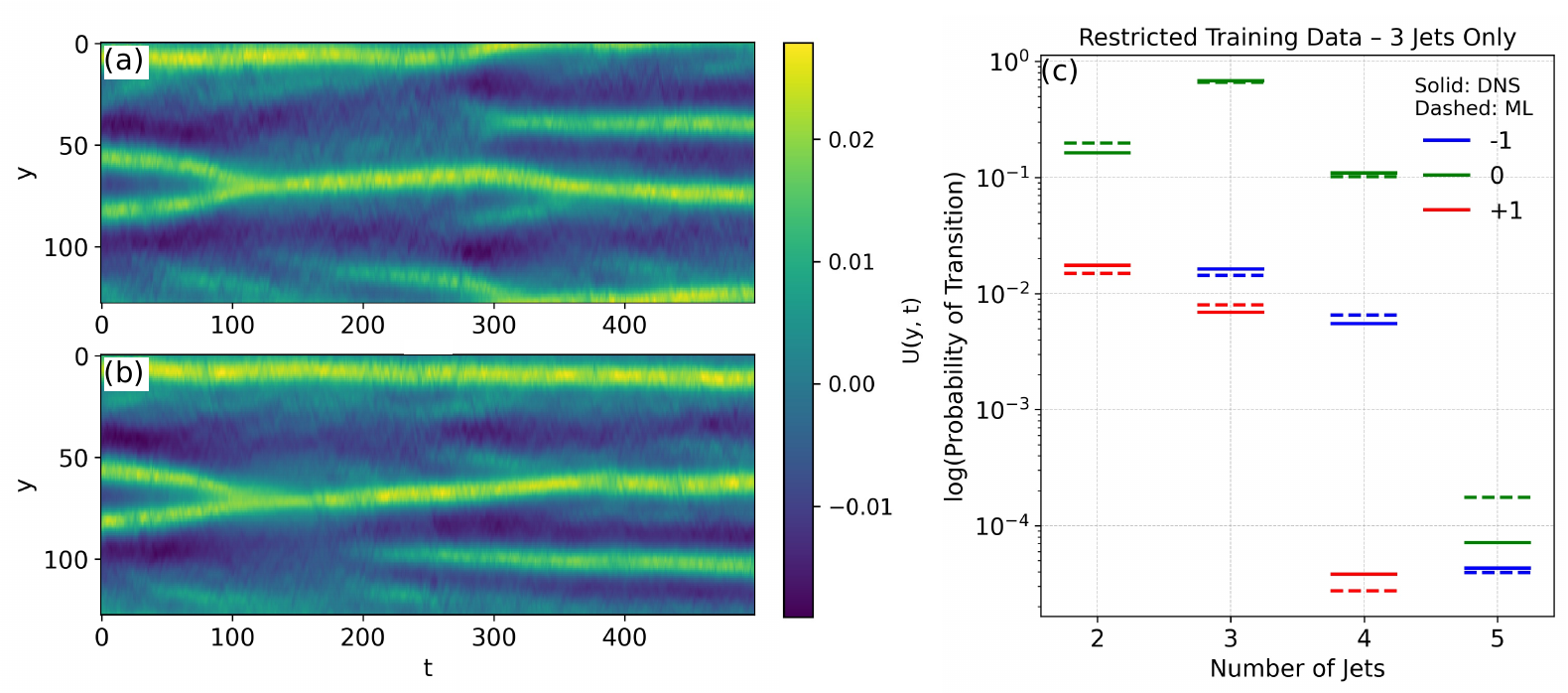}
\caption{Prediction of out-of-training-distribution dynamics for beta-plane turbulence, where the training dataset includes only states with 3 jets.
(a) Zonally-averaged zonal velocity $U_1$ from the upper layer, obtained via DNS.
(b) $\tilde{U_1}$ predicted by the NN from the same initial condition as (a), with the network trained on a dataset including only 3 jets states, with no transition events.
(c) Transition probabilities for the number of jets, as in Fig. 2, comparing DNS (solid lines) for the full system and the NN trained on the restricted dataset (dashed lines).
}
\end{figure}

We now consider how our emulator captures the dynamics of a second system: two-layer beta-plane turbulence, a well-studied model in geophysical fluid dynamics known for its jet-like structures—localised regions of high velocity (yellow in Fig. 4). The system is governed by the quasi-geostrophic (QG) potential vorticity equations:
\[
\partial_t \zeta_m + J(\psi_m, \zeta_m) + \beta \partial_x \psi_m + U_m \partial_x \zeta_m = \delta_{m,2} \mu \nabla^2 \psi_2 + \text{ssd},
\]
where the vorticity in each layer \( m = \{1,2\} \) is given by $\zeta_m = \nabla^2 \psi_m + (-1)^m k_d^2 (\psi_2 - \psi_1) \in R^{128 \times 128}$. The deformation wavenumber is $k_d^2 = \frac{f_0^2}{g'}\frac{H_1+H_2}{H_1 H_2}$, with \( H = H_1 + H_2 \) being the total depth at rest. Here, \( f_0 \) is the Coriolis parameter at a reference latitude, and $g'$ is the reduced gravity, accounting for stratification, and $J(\psi_m, \zeta_m):=\partial_x \psi_m \partial_y \zeta_m - \partial_y \psi_m \partial_x \zeta_m$. The parameter \( \beta \) represents the planetary vorticity gradient, while $U_m(y,t) = \frac{1}{L_x} \int_0^{L_x} -\partial_y \psi_m \, dx$ denotes the zonally-averaged (\( x \)-mean) zonal velocity in layer \( m \). The drag coefficient \( \mu \) applies to the bottom layer (\( m = 2 \)), and small-scale dissipation (\(\text{ssd}\)) is implemented via an exponential filter (see End Matter).  

DNS was performed on a CPU using PYQG \cite{pyqg} in Python \cite{python} using a pseudo-spectral method with a third-order Adams-Bashforth scheme with $\delta t = 1$ hour downsampled to $\Delta t= 10$ days for training, initialised with small amplitude noise in $\zeta$ from a uniform distribution over the interval $[-10^{-6}, 10^{-6}]$. Again only dynamics after a warm-up phase was used as training data. Previous work has demonstrated successful NN emulation across varying $\beta$ for the 1-layer stochastically forced beta-plane equation \cite{shokar2024extending}. Here, we fix $\beta = 5 \times 10^{-11} \, \text{s}^{-1}$ (with other parameters detailed in the End Matter) and construct a restricted dataset containing only three jet states, with no transitions to higher or lower jet counts. This restricted dataset is then used to train the emulator.

Given initial conditions $\psi_m$ for both layers, the NN generates an evolution autoregressively, as with the KS equation. For visualisation of the jets that form, we plot the (upper layer) zonally-averaged zonal velocity $U_1$ over time. Fig. 4(a) shows the evolution generated by DNS, while Fig. 4(b) demonstrates that the NN successfully captures key transition events, including coalescence (~100 days) and nucleation (~200 days), before diverging due to the system's intrinsic unpredictability—despite being trained solely on states with three jets and no transitions. The transition probabilities shown in Fig.4 (b) further confirm that, similar to the KS equation, the NN is capable of forecasting dynamics beyond its training dataset.

%

%
In summary, we have shown using two archetypal chaotic PDEs that ML emulators can predict unobserved dynamics at the same parameter values as the training data (by capturing relaminarisation, initialisation and 'kink' transitions in the KS equation and jet transitions in beta-plane turbulence) and also successfully extrapolate outside the training set parameter regime (by doubling $L$ in the KS equation). Despite employing relatively generic architectures (a transformer with local attention), our emulators are still able to challenge the notion that ML cannot generalise beyond its training data.
%
%

We conjecture that this success stems from the model learning the discrete evolution operator $\mathcal{F}_{\Delta t}$ defined by the PDE $u_t = F(u)$ sufficiently well over $u$  rather than merely 'memorising' and interpolating observed patterns seen in training. The training set need not be large but must sufficiently span phase space for $\mathcal{F}_{\Delta t}$ to be learned effectively, noting that training on a local region in physical space has been shown to fail \cite{Scher}. An important point to make here is that the functional ${\cal F}_{\Delta t}$ being approximated can be, and probably typically is, far simpler than the dynamics (or patterns) it produces: the logistic map being an exemplar.

One possible reason for our success, where other methods have struggled, may be the optimisation approach itself. Unlike physics-informed neural networks (PINNs), which impose explicit equation constraints and often face challenging optimisation landscapes, our approach learns directly from data, enabling more stable training and a higher likelihood of reaching a global optimum instead of getting trapped in local minima. There is, of course, the further important ingredient that the number of degrees of freedom in the systems is relatively small and the data is noise-free, however we believe there is no conceptual issue with this being scaled.
%
%

In climate science, capturing extreme behaviour is crucial for accurate forecasting.  However, typically uncertainty in the PDE model is compensated by added stochasticity. This poses additional challenges for ML approaches. Despite this, preliminary results using the approaches outlined here with a probabilistic ML emulator suggest promising potential, not only for capturing previously unseen dynamics from incomplete information but also for quantifying the uncertainty associated with their occurrence.

\textit{Acknowledgments---}I.S. acknowledges funding by the UK Engineering and Physical Sciences Research Council (grant number EP/S022961/1) as part of the UKRI Centre for Doctoral Training in Application of Artificial Intelligence to the Study of Environmental Risks.

\bibliographystyle{ieeetr}
\bibliography{references}

\newpage

\appendix

\section{Local Attention}

Local attention focuses on a spatially localized window around each point, akin to the receptive field in convolutional networks. It captures small-scale interactions, while global structures are handled across layers. By leveraging attention mechanisms instead of traditional convolutional operations \cite{CNN}, this approach dynamically computes spatial weights, ensuring adaptability to varying input scales. 

The input $u \in \mathbb{R}^D$, where $D$ represents the number of spatial grid points, is encoded into a latent space $z$ for transformer operations. This encoding is performed via a linear transformation with weights $W_{\text{E}} \in \mathbb{R}^{1 \times C}$, where $C$ denotes the number of channels. After processing through the transformer blocks, a decoder applies a linear transformation that maps the $C$ channels back to a single channel, yielding the prediction $\tilde{u}_{t+1} \in \mathbb{R}^D$. The overall architecture is schematically depicted in Fig.5.

\begin{figure}[t!] 
\centering
    \includegraphics[width=0.5\textwidth]{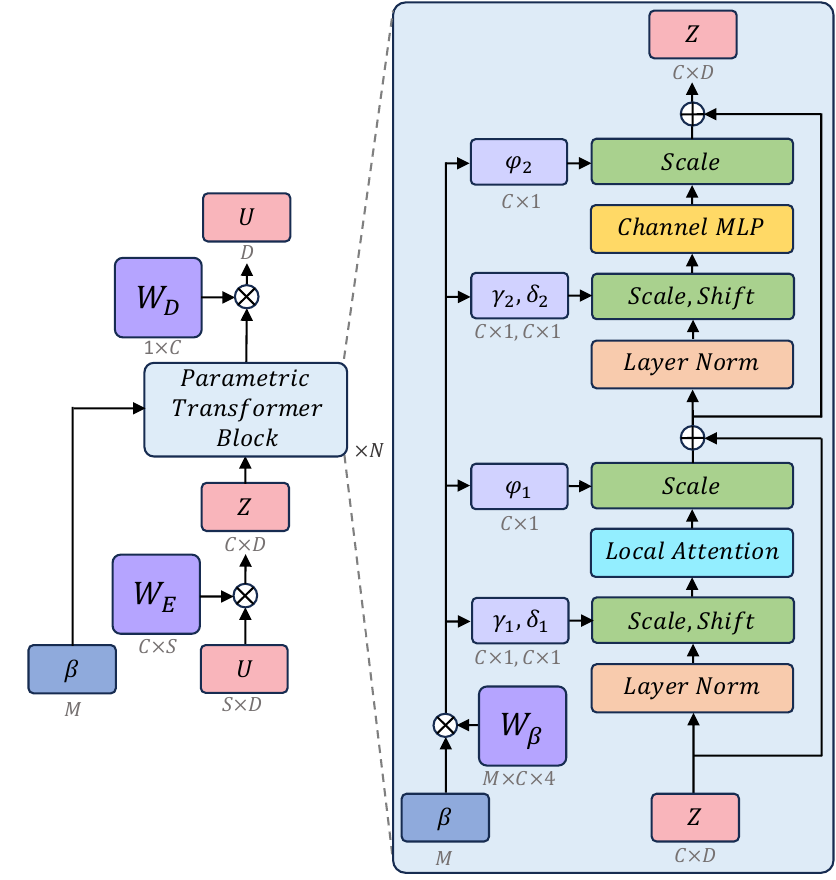}
        \caption{
       Schematic of the NN architecture. The network is structured around a transformer architecture conditioned on parameter $L$ and initialised by conditions $U$. Within each transformer block, adaptive layer normalisation conditions the transformer on $L$ by replacing scale and shift parameters. Each $W$ denotes learned weights for linear transformations, with arrows indicating the forward pass. In this study, the conditioning parameter $L$ is defined with a size of $M=1$ and the temporal history provided to the model $S=1$. However, the architecture is flexible to extension to larger dimensions within the parameter space.
        }
    \end{figure} 

The local attention layer computes correlations within a neighbourhood $\mathit{N}_K(x)$ of spatial extent $K$ around each position $x$ in $z \in \mathbb{R}^{C \times D}$. For $x$, attention is computed as $a_x = \sum_{b \in \mathit{N}_K(x)} \text{softmax}_b(\zeta_x^T k_b) v_b$, where $q$, $k$, and $v \in \mathbb{R}^{C \times D}$ are query, key, and value vectors from linear projections of $z$ \cite{Attention}. This mechanism extracts local contextual information, enhancing focus on relevant neighbourhood features. In our implementation, we also employ circular padding to ensure that the neighbourhoods wrap around at the boundaries of the input tensor, to retain translational equivariance. To encode relative positional information between elements, we introduce a learnable relative positional encoding, $\text{PE} \in \mathbb{R}^{K}$. This approach achieves a large receptive field with fewer layers, avoiding pooling or striding. It is especially beneficial for multi-scale dynamics. By using relative positions and circular padding in the unfold operation, the attention mechanism preserves translation equivariance, aligning with the symmetries of the dynamics.

\subsection{Parametric Transformer}

\begin{figure}[t!] 
\centering
    \includegraphics[width=\textwidth]{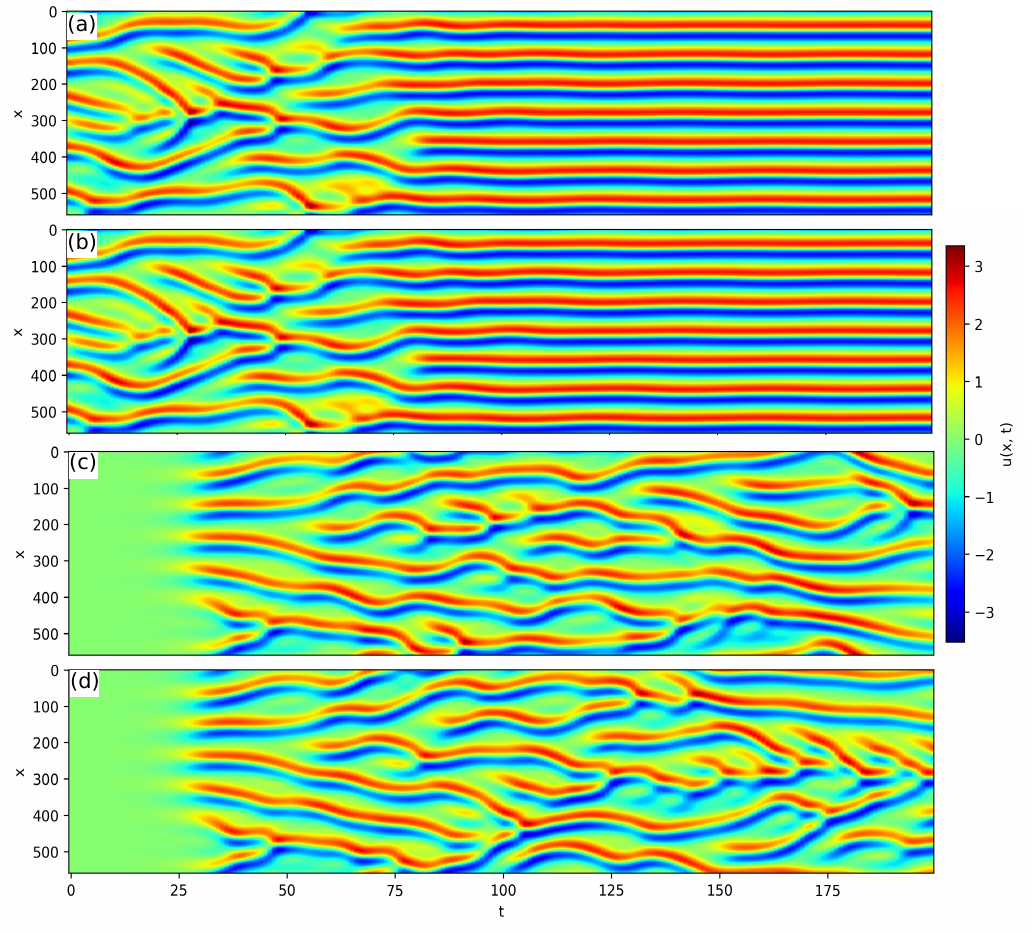}
        \caption{(a) Relaminarisation event from Fig.1 (a) captured by a single layer fully-connected NN trained and evaluated on $L=56$, where no laminar dynamics were present in the training data. (b) Same as (a) generated by FNO. (c) Initialisation of KS flow from initial conditions from Fig.1 (d) emulated by a single layer MLP trained and evaluated on $L=56$. (d) Same as (c) for the FNO.}
\end{figure} 

\begin{figure*}[t!] 
    \centering
    \includegraphics[width=\textwidth]{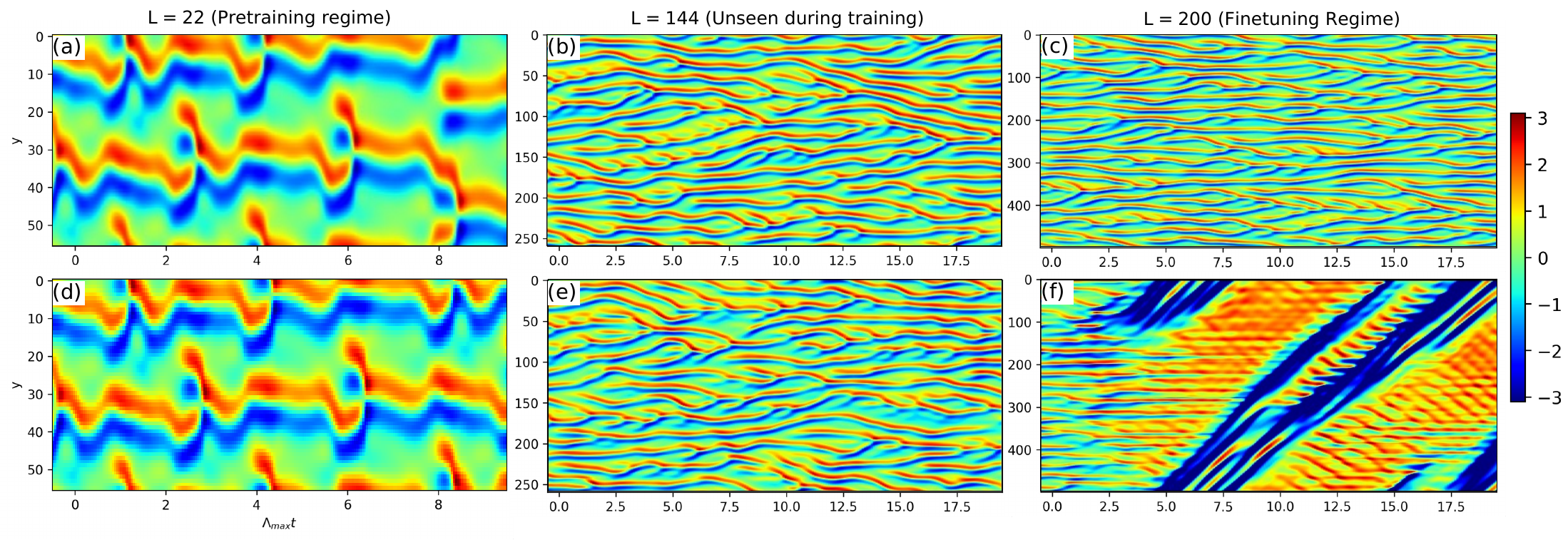}
    \caption{Generalisation capability of the Neural Network. All are pretrained with $L=22$ before finetuning on $L=\{22, 36, 48, 64, 98, 128, 200\}$. (a.-c.) Local Attention based NN. (d.-f.) FNO.}
\end{figure*}

We replace the affine parameters in layer normalisation with a learned function of the conditioning information \cite{FILM}. This approach enables leaner pre-training by decoupling scale and shift parameters from the conditioning variables, improving transformations during fine-tuning. We define scaling ($\gamma$) and shift ($\delta$) parameters to modulate activations in both the attention operation and MLP, totalling four parameters. These are obtained via a linear transformation $W_\beta \in \mathbb{R}^{M \times 4C}$ applied to the context vectors $\beta$, where $M=1$ represents their size. While this work focuses on $M=1$, the framework is easily extendable to higher-dimensional parameter spaces. Transformations within each transformer block are given by:

\begin{align}
    z &\rightarrow z + \varphi_1 \cdot \mathrm{LA}\left(\gamma_1 \cdot \mathrm{LN}(z) + \delta_1\right), \\
    z &\rightarrow z + \varphi_2 \cdot \mathrm{MLP}\left(\gamma_2 \cdot \mathrm{LN}(z) + \delta_2\right),
\end{align}

Here, $z$ is the transformer's hidden state, $\mathrm{LA}$ denotes local attention, $\mathrm{LN}$ represents layer normalisation, and $\mathrm{MLP}$ is a two-layer perceptron with a GELU \cite{gelu} activation between the layers.

\section{Kuramoto-Sivashinsky Equation}

As described in the main text, the emulators trained on the restrictive datasets shown in Fig. 2 are capable of both forecasting relaminarisation and initialising flows for the KS equation, as demonstrated in Fig. 6. In both cases, we observe transitions between states with $n$ kinks to $n+1$ and $n-1$ kinks and the emulators successfully reproduce these transitions, despite not encountering them during training.

Fig. 7(f) highlights the limitations of the Fourier Neural Operator (FNO) in generalising across domains of varying sizes. While the FNO generates realistic forecasts for domains of size $L=22$ and $L=144$ in Figures Fig. 7(d) and Fig. 7(e), it struggles to produce a plausible forecast when the domain size increases to $L=200$, as shown in Fig. 7(f) This limitation arises because the FNO explicitly models only the largest $N$ wavenumbers, relying on a $1 \times 1$ convolution to recover higher wavenumbers. For example, with $L=22$ and $N_x=56$ resolved grid points, only the highest $N=28$ wavenumbers are modelled by the FNO. When $L=200$, the system has approximately 31 unstable modes (estimated as $N_{\text{unstable}} \approx \frac{L}{2 \pi}$), which exceeds the capacity of an FNO trained on $L=22$ to capture all dynamically unstable modes at the larger domain size. In contrast, the local attention mechanism is designed to model local dynamics explicitly by learning a translation invariant stencil over the domain, allowing it to flexibly adapt to domains of arbitrary size without sacrificing the ability to resolve dynamics across scales.

\section{Beta-Plane Turbulence}

DNS of the two-layer beta-plane equation was performed using PYQG \cite{pyqg} in Python \cite{python} with the following parameters: $\beta = 5 \times 10^{-11} \, \text{s}^{-1}$, bottom layer damping $\mu = 4 \times 10^{-8} \, \text{s}^{-1}$, Rossby radius of deformation $R_d = 12.5 \, \text{km}$, depth of the upper layer $H_1 = 500 \text{m}$, $H_2 = 1500 \text{m}$ and the upper layer flow speed is $U_1 = 0.025 \text{m} \text{s}^{-1}$.

The small-scale dissipation (ssd) removes enstrophy cascading to small scales, preventing energy buildup at the grid scale that could cause numerical instability. Acting as a hyperviscosity-like term, it selectively damps high-wavenumber modes while preserving large-scale dynamics.  In Fourier space, this is applied via an exponential filter:  

\[
E_f = 
\begin{cases} 
\exp\left[C_{\text{ssd}}\,(\kappa^{\star} - \kappa_c)^4\right], & \quad \kappa \geq \kappa_c, \\ 
1, & \quad \text{otherwise}.
\end{cases}
\]

where $\kappa^{\star} = \sqrt{ (k\,\Delta x)^2 + (l\,\Delta y)^2 }$ is a non-dimensional wavenumber and \(\kappa_c\) is set to 65\% of the Nyquist scale, \((\kappa^{\star}_{\text{ny}} = \pi)\).  The constant \(C_{\text{ssd}}\) ensures energy at \(\kappa^{\star} = \pi\) vanishes within machine precision C$_{\text{ssd}} = \frac{\log 10^{-15}}{(0.35\, \pi)^4} \approx -23.5$.

\end{spacing}
\end{document}